\documentclass[12pt, preprint]{elsarticle}

\usepackage{amssymb}

\usepackage{lineno}
\usepackage{multirow}
\usepackage{url}   
\usepackage{subfigure}

\usepackage{color}
\definecolor{lgrey}{rgb}{0.4 ,0.4,0.4}
\definecolor{hgrey}{rgb}{0.15 ,0.15,0.15}

\usepackage[utf8]{inputenc}

\usepackage{listings}
\usepackage{pdfpages}
\usepackage{graphics}
\usepackage{array}
\definecolor{gray}{rgb}{0.3 ,0.3,0.3}
\definecolor{cadmiumgreen}{rgb}{0.0, 0.42, 0.24}

\usepackage[usenames,dvipsnames,svgnames,table]{xcolor}

\definecolor{lightColor}{HTML}{d3dfd3}
\definecolor{mediumColor}{HTML}{92ae90} 
\definecolor{darkColor}{HTML}{507d4d}
\definecolor{chapterColor}{HTML}{507d4d}
\definecolor{linkColor}{HTML}{7a4d7d}
\definecolor{colbackColor}{HTML}{ded3df}

\usepackage{listings}
\definecolor{gray}{rgb}{0.4,0.4,0.4}
\definecolor{darkblue}{rgb}{0.0,0.0,0.6}
\definecolor{cyan}{rgb}{0.0,0.6,0.6}
\lstdefinestyle{XML}
{
  basicstyle=\footnotesize\ttfamily,
  morestring=[b]",
  morestring=[s]{>}{<},
  morecomment=[s]{<?}{?>},
  stringstyle=\color{black},
  identifierstyle=\color{cyan},
  keywordstyle=\color{darkblue},
  showtabs=false,                  
  tabsize=2,
  morekeywords={header, configuration, classifier, multiInstanceClassifier, evaluator, data,   trainFile,testFile,xmlFile,reportFileName, multiLabelClassifier, parameters, parameter, value, class, transformMethod, standardDeviation, macroMeasuresLabels, threshold, seed, sampleWithReplacement, useConfidences, samplePercentage, numClassifiers, baseLearner, measures, measure, numFolds, nReferences, nCiters, metric, listOptions, report, fileName, @attribute, @relation, @data, label, @end, labels, file}
}

\lstdefinestyle{XMLInline}
{
  basicstyle=\small\ttfamily,
  morestring=[b]",
  morestring=[s]{>}{<},
  morecomment=[s]{<?}{?>},
  stringstyle=\color{black},
  identifierstyle=\color{cyan},
  keywordstyle=\color{darkblue},
  showtabs=false,                  
  tabsize=2,
  morekeywords={header, configuration, classifier, multiInstanceClassifier, evaluator, data,   trainFile,testFile,xmlFile,reportFileName, multiLabelClassifier, parameters, parameter, value, class, transformMethod, standardDeviation, macroMeasuresLabels, threshold, seed, sampleWithReplacement, useConfidences, samplePercentage, numClassifiers, baseLearner, measures, measure, numFolds, nReferences, nCiters, metric, listOptions, report, fileName, @attribute, @relation, @data, label, @end, labels, file}
}

\definecolor{pblue}{rgb}{0.13,0.13,1}
\definecolor{pgreen}{rgb}{0,0.5,0}
\definecolor{pred}{rgb}{0.9,0,0}
\definecolor{pgrey}{rgb}{0.46,0.45,0.48}

\lstdefinestyle{numbers} {numbers=left, stepnumber=1, numberstyle=\tiny, numbersep=10pt}

\lstdefinestyle{Java}
{language=Java,
commentstyle=\color{cadmiumgreen},
  keywordstyle=\color{blue},
  stringstyle=\color{magenta},
  xleftmargin=9pt,
  xrightmargin=2pt,
}
\usepackage{xfp} 
\newcommand{\toMLAlgorithms}{15}
\newcommand{\toMIAlgorithms}{15}
\newcommand{\MIMLAlgorithms}{13}

\usepackage{hyperref} 
\journal{Neurocomputing}

\begin{document}

\begin{frontmatter}




\author{Álvaro Belmonte\fnref{label1}}
\ead{i32bepea@uco.es}
\author{Amelia Zafra\fnref{label1,label2}}
\ead{azafra@uco.es}
\author{Eva Gibaja\fnref{label1,label2}\corref{cor1}}
\ead{egibaja@uco.es}
\cortext[cor1]{Corresponding author.}
\address[label1]{Department of Computer Science and Numerical Analysis. University of Córdoba, Spain}
\address[label2]{Andalusian Research Institute in Data Science and  Computational Intelligence (DaSCI). University of Córdoba, Spain}

\title{MIML library: a Modular and Flexible Library for Multi-instance Multi-label Learning}


\begin{abstract}
MIML library is a Java software tool to develop, test, and compare classification algorithms for multi-instance multi-label (MIML) learning. The library includes \inteval{\toMLAlgorithms+\toMIAlgorithms+\MIMLAlgorithms} algorithms and provides a specific format and facilities for data managing and partitioning, holdout and cross-validation methods, standard metrics for performance evaluation, and generation of reports. In addition, algorithms can be executed through $xml$ configuration files without needing to program. It is platform-independent, extensible, free, open-source, and available on GitHub under the GNU General Public License.

\end{abstract}


\end{frontmatter}




\section{Introduction}\label{sec:introduction}

Multi-instance multi-label (MIML) learning has emerged as a promising supervised learning paradigm by combining multi-instance (MI) and multi-label (ML) learning. On the one hand, MI allows a more flexible representation of input space by associating a pattern (\emph{bag}) with multiple instances, all of them with the same number of attributes~\citep{Die1997}. On the other hand, ML introduces greater flexibility representation in the output space by associating a pattern with not one, but a set of binary classes (\emph{labels})~\citep{Gib2015tutorial}. For example, an image could be represented by multiple instances being each one a region in the image and each image could have several binary labels (e.g. \textsl{cloud}, \textsl{lion}, \textsl{landscape}). MIML carries out a natural formulation of complex objects in real problems such as text and image categorization, audio and video annotation, or bioinformatics~\citep{Her2016}. 

Currently, there are available several well-known Java frameworks such as Weka~\citep{Hal2009} to work with MI learning and Mulan~\citep{Tsoumakas:2011:MJL:1953048.2021078} or Meka~\citep{read2016meka} to work with ML learning. Nevertheless, none of these libraries can work with MIML data. To the best of our knowledge, there are only publicly available some Matlab implementations of particular MIML algorithms~\citep{lamda}. However, they are not integrated into a library. Each one has its specific configuration and data format.
~The motivation of this work is to develop a modular and extensible Java library, based on the Weka and Mulan libraries, to facilitate the running and development of MIML classification algorithms.

The rest of the document is organized as follows: an overview of the architecture and functionalities of the library is specified in Section \ref{sec:softwareFramework}. An illustrative example is presented in Section \ref{sec:illustrativeExamples}. Finally, conclusions are pointed out in Section \ref{sec:conclusion}.

\section{MIML Framework}\label{sec:softwareFramework}

\subsection{Software Architecture}
\begin{figure}[!h]
  \includegraphics[width=0.98\textwidth]{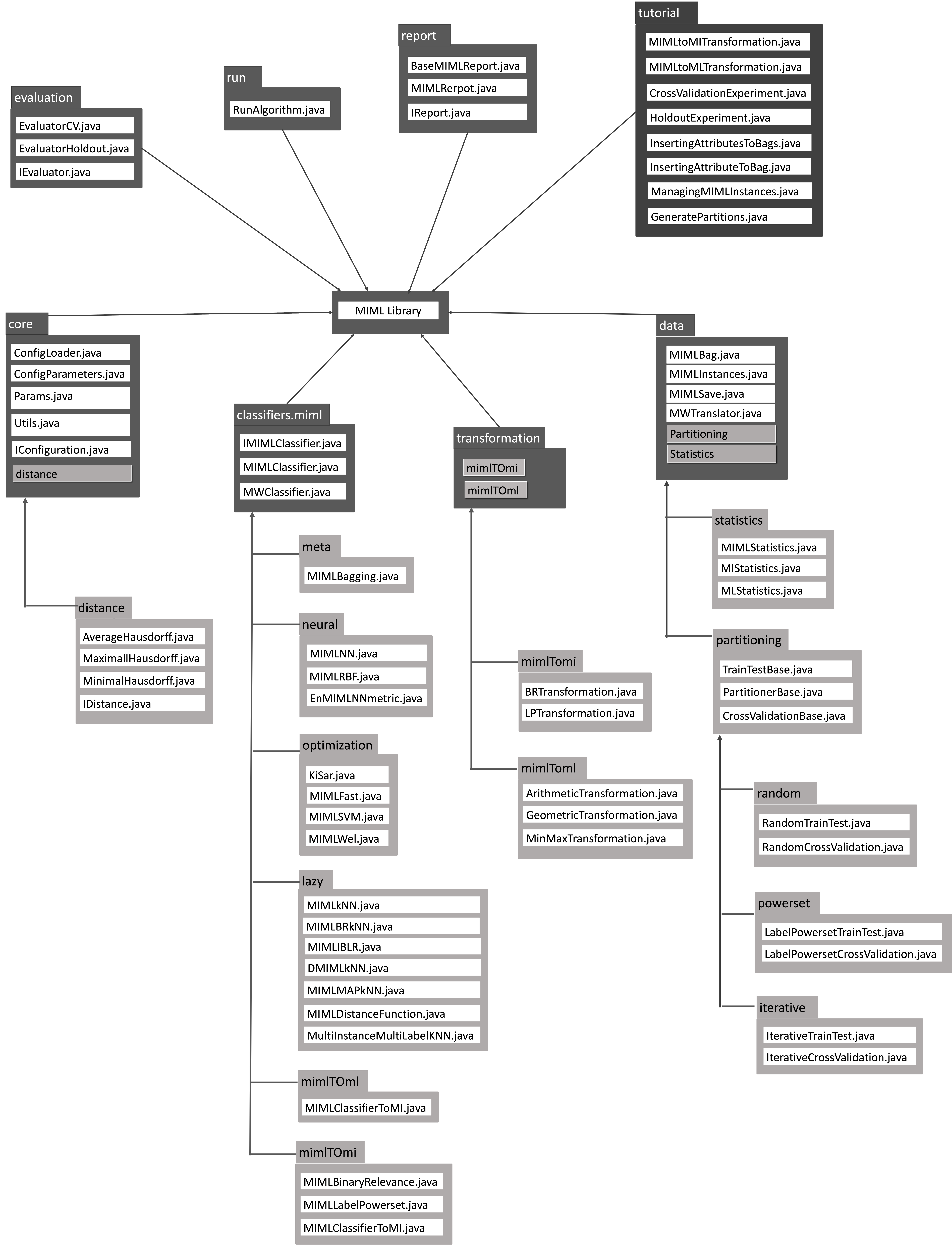}
  \caption{Architecture of MIML library.}
  \label{fig:arquitectura}
\end{figure}

MIML library is available in GitHub\footnote{\url{https://github.com/kdis-lab/MIML}} under the GNU license. It is organized in packages, each one containing the interfaces and the classes required to extend its functionality (see Figure~\ref{fig:arquitectura}).
The \textit{core} and \textit{run} packages contain classes related to the configuration and running of algorithms by means of \textit{xml} files. The \textit{evaluation} package contains cross-validation and holdout evaluators.
The \textit{data} package includes classes to load and save MIML data files and two subpackages: \textit{miml.data.statistics} with classes to obtain statistics about the dataset and \textit{miml.data.partitioning} with classes for holdout and cross-validation data partitioning. ~The \textit{transformation} package comprises methods to transform a MIML dataset into either ML (\textit{transformation.mimlTOml}) or MI dataset (\textit{transformation.mimlTOmi}).
Interfaces and abstract classes required to develop MIML classification algorithms are included in \textit{classifers.miml} which comprises six subpackages: \textit{mimlTOmi} and \textit{mimlTOml} with algorithms which transform the MIML problem to a MI or ML problem respectively. The \textit{meta} package is for ensemble methods, \textit{lazy} for instance-based algorithms, \textit{neural} for neural networks algorithms, and \textit{optimization} for proposals based on optimization.
The \textit{report} package contains classes to generate reports for the experiments carried out, and the \textit{tutorial} package includes practical usage examples (e.g. transforming and managing a dataset, running holdout and cross-validation experiments, data partitioning, etc.).
In addition, the distribution provides an extensive user manual with a complete description of the library\footnote{\url{https://github.com/kdis-lab/MIML/tree/master/documentation}}. It includes topics such as an introduction to MIML learning, getting and running the library, managing MIML data, running a classification algorithm included in the library, developing a new MIML classification algorithm and practical examples for all algorithms. The API Javadoc in both \textit{html} and \textit{pdf} is also provided. 

\subsection{Software Functionalities}
The library uses a data format specifically designed for MIML learning which is based on Weka and Mulan data formats. It additionally provides functionality for loading, manipulating, saving, and obtaining representative metrics of MIML data (e.g. number of bags, number of labels, cardinality, density, label co-occurrence, imbalance ratio, etc.). Besides, 3 strategies for holdout and cross-validation MIML data partitioning are available: random, powerset~\cite {Sec2011} and iterative~\cite {Sec2011}. The library includes 3 transformations of MIML data to ML representation (arithmetic, geometric, and min-max) and 2 transformations to MI representation (BR and LP) as well as 3 different distance functions between bags (average, minimal, and maximal Hausdorff). It can run both cross-validation and holdout experiments.

 The library includes \inteval{\toMLAlgorithms+\toMIAlgorithms+\MIMLAlgorithms} MIML classification algorithms (see Table~\ref{tbl:algorithms}). On the one hand, $\toMIAlgorithms$ MI Weka algorithms can be used by transforming the problem to MI. On the other hand, $\toMLAlgorithms$ Mulan algorithms can be executed by transforming the problem to ML. Besides, the library includes $\MIMLAlgorithms$ algorithms that cope with a MIML problem without transformation, some of them developed as wrappers. 
 Concretely, MIMLBagging that follows a bagging scheme~\citep{breiman1996bagging}, algorithms based on neural networks such as MIMLNN~\citep{ZHOU20122291}, MIMLRBF~\citep{zha2009mimlrbf} and EnMIMLNNmetric~\citep{wu2014genome}; algorithms based on optimization such as MIMLFast~\citep{Huang2019}, KiSar~\citep{Li2012Towards}, MIMLSVM~\citep{Zhou2006} and
MIMLWel~\citep{Yang2013multi} and algorithms based on nearest neighbors such as MIMLkNN algorithm~\citep{Zha2010}. Finally, it includes MIMLBRkNN, MIMLMAPkNN, DMIMLkNN and MIMLIBLR that are adaptations to the MIML framework of BRkNN\_ML~\cite{Ele2008}, MLkNN~\cite{Zha2007}, DMLkNN~\citep{You2008} and IBLR\_ML~\cite{Che2009} ML algorithms. 

 Algorithms can be easily executed through configuration files in \textit{xml} format to set holdout and cross-validation methods and a wide set of performance metrics that will include the output report. In addition to this easy way to run the algorithms, its modular structure eases the development of new proposals simply extending the class \textit{miml.classifiers.miml.MIMLClassifier}.

\begin{table}
\caption{Algorithms included in the MIML library.} 
\centering
\scriptsize
\begin{tabular}{c|l||c|l||c|l}
\rowcolor{gray}\multicolumn{2}{c||}{\textcolor{white}{\textbf{MIMLtoMI}}}&\multicolumn{2}{c||}{\textcolor{white}{\textbf{MIMLtoML}}}&
\multicolumn{2}{c}{\textcolor{white}{\textbf{MIML}}}\\
\hline

\rowcolor{lightgray}
\textbf{Label}&\textbf{MI Algorithm}&\textbf{Bag}&\textbf{ML Algorithm}&\multicolumn{2}{c}{\textbf{MIML}}\\ 

\rowcolor{lightgray}
\textbf{transf.}&\textbf{(Weka)}&\textbf{transf.}&\textbf{(Mulan)}&\multicolumn{2}{c}{\textbf{Algorithm}}\\
\hline
&CitationKNN~\citep{Wang2000}&&BR~\citep{tsoumakas2009}&&\\
&MDD~\citep{Maron1998Framework}&&LP~\citep{tsoumakas2009}&Bagging&MIMLBagging\\
\cline{5-6}
&MIDD~\citep{Maron1998Framework}&&RPC~\citep{Hul2008}&&MIMLNN~\citep{ZHOU20122291}\\
&MIBoost~\citep{Xu2004}&&CLR~\citep{Fur2008}&ANN&MIMLRBF~\citep{zha2009mimlrbf}\\
&MILR~\citep{ray2005supervised}&&BRkNN~\citep{Ele2008}&&EnMIMLNNmetric~\citep{wu2014genome}\\
\cline{5-6}
BR~\citep{tsoumakas2009}&MIOptimalBall~\citep{Auer2004}&&DMLkNN~\citep{You2008}&&MIMLkNN~\citep{Zha2010}\\
&MIRI~\citep{Bjerring2011}&Arithmetic&IBLR\_ML~\citep{Che2009}&&DMIMLkNN
\\
&MISMO~\citep{Gartner2002}&Geometric&MLkNN~\citep{Zha2007}&Lazy&MIMLIBLR
\\
&MISVM~\citep{Andrews2003}&Min-Max&HOMER~\citep{Tsoumakas2008}&&MIMLMAPkNN
\\
&MITI~\citep{Blockeel2005}&~\citep{Dong2006}&RAkEL~\citep{Tso2011}&&MIMLBRkNN
\\
\cline{5-6}
&MIWrapper~\citep{Frank2003}&&PS~\citep{Rea2008}&&MIMLWel~\citep{Yang2013multi}\\
&SimpleMI~\citep{Dong2006}&&EPS~\citep{Rea2008}&Other&MIMLFast~\citep{Huang2019} \\
\cline{1-2}
&CitationKNN~\citep{Wang2000}&&CC~\citep{Rea2011}&&KiSar~\citep{Li2012Towards}\\
LP~\citep{tsoumakas2009}& MIWrapper~\citep{Frank2003}&&ECC~\citep{Rea2011}&&MIMLSVM~\citep{Zhou2006}\\
& SimpleMI~\citep{Dong2006}&&MLStacking~\citep{GrigoriosDimou2009}&&\\
\hline
\end{tabular}

\label{tbl:algorithms}
\end{table}

\section{An illustrative example}\label{sec:illustrativeExamples}

Classification algorithms included in the library are executed by the \textit{RunAlgorithm} class by using a \textit{xml} configuration file. The configuration file path is specified through the command line with the option \textit{-c}. An example of an execution command would be: 

\begin{scriptsize}
\begin{lstlisting}[language=bash]
  $ java -cp dist/miml-1.0.jar  miml.run.RunAlgorithm -c dist/configurations/MIMLClassifier/MIMLkNN.config
\end{lstlisting}
\end{scriptsize}

Configuration files start at root element \lstinline[style=xmlinline]{<configuration>} and contain three branches: \lstinline[style=xmlinline]{<classifier>}, \lstinline[style=xmlinline]{<evaluator>} and \lstinline[style=xmlinline]{<report>}. The \lstinline[style=xmlinline]{<classifier>} element specifies the classification algorithm by means of its \lstinline[style=xmlinline]{name} attribute. 
In the example, the specification of the parameters \lstinline[style=xmlinline]{<nReferences>}, \lstinline[style=xmlinline]{<nCiters>} and \lstinline[style=xmlinline]{<metric>} of MIMLkNN algorithm is shown.
The \lstinline[style=xmlinline]{<evaluator>} element sets the validation method by means of its
\lstinline[style=xmlinline]{name} attribute. The dataset used is described in element \lstinline[style=xmlinline]{<data>}. 
In the example, cross-validation evaluator is used to set the seed and the number of folds in \lstinline[style=xmlinline]{<seed>} and \lstinline[style=xmlinline]{<numFolds>} elements.
Finally, the \lstinline[style=xmlinline]{<report>} element specifies the class of the output report by means of \lstinline[style=xmlinline]{name} attribute. This class can be easily extended to obtain the most convenient output format.
The \lstinline[style=xmlinline]{<fileName>} element  contains the path where the report will be stored. Optionally, the \lstinline[style=xmlinline]{<measures>} element can be defined to include a specific set of measures in the output report. If no measures are configured, all measures allowed by the classifier are shown. In the example, the \textit{hamming loss}, \textit{macro-averaged precision} and \textit{micro-averaged recall} are specified. For macro-averaged measures, the \lstinline[style=xmlinline]{perLabel} attribute sets if separated measures for each label are included in the report. Examples of configuration files for all algorithms in the library are included in \textit{configurations} folder as well as in the user's manual. The user manual details the rest of the functionalities (e.g. transforming a dataset, developing new proposals, etc.).

\begin{lstlisting}[style=XML]
<classifier name="miml.classifiers.miml.lazy.MIMLkNN">
	<nReferences>4</nReferences>
	<nCiters>6</nCiters>
	<metric name="miml.core.distance.AverageHausdorff"></metric>
</classifier>
<evaluator name="miml.evaluation.EvaluatorCV">
    <seed>712637</seed>
	<numFolds>5</numFolds>
	<data>
		<file>data/miml_birds.arff</file>
		<xmlFile>data/miml_birds.xml</xmlFile>
	</data>
</evaluator>
<report name="miml.report.BaseMIMLReport">
	<fileName>results/mimlknnn.csv</fileName>
	<measures perLabel="true">
		<measure>Hamming Loss</measure>
		<measure>Macro-averaged Precision</measure>
		<measure>Micro-averaged Recall</measure>
	</measures>
</report>
\end{lstlisting}

\section{Conclusion}\label{sec:conclusion}
This work presents a free, open-source Java library that eases the development, testing, and comparison of MIML classification algorithms. As it is based on Weka and Mulan, researchers on MIML who have used any of these libraries will be familiar with it. Besides, it is extensible, platform-independent, and easy to install, configure and use. It is accompanied by a user manual with usage examples, a step-by-step tutorial on preparing and running experiments, a detailed overview of the architecture and instructions to extend the software for customization purposes. We consider this library can facilitate the development of experimental studies and the design of new proposals, contributing to future advances in the field.

\section*{Acknowledgments}
This research was supported by the Spanish Ministry of Science and Innovation and the European Regional Development Fund, project PID2020-115832GB-I00.

\label{sec:references}

\begin{scriptsize}
  \bibliographystyle{elsarticle-num} 
  \bibliography{bibliography}
\end{scriptsize}





\clearpage
\section*{Required Metadata}

\section*{Current executable software version}

\small


\begin{table}[!htp]
\caption{Software metadata}
\centering\scriptsize 
\begin{tabular}{| p{0.5cm} | p{5.5cm} | p{8cm} |} \hline
\textbf{Nr.} & \textbf{(executable) Software metadata description} & \textbf{Please fill in this column} \\ \hline
S1  & Current software version                      & 1.0 \\ \hline
S2  & Permanent link to executables of this version & https://github.com/kdis-lab/miml \\ \hline
S3  & Legal Software License                        & GPLv3 \\ \hline
S4  & Computing  platform/Operating System          & GNU Linux, Microsoft Windows, OS X \\ \hline
S5  & Installation requirements \& dependencies     & Java version 1.8\\ \hline
S6  & If available, link to user manual - if formally published include a reference to the publication in the reference list &  https://github.com/kdis-lab/MIML/blob/master/ documentation/MIML-UserManual.pdf \\ \hline
S7 & Support email for questions  & azafra@uco.es \\ \hline
\end{tabular}

\label{table:softwaremetadata}
\end{table}

\section*{Current code version}

\begin{table}[!htp]
\caption{Software metadata}
\centering\scriptsize 
\begin{tabular}{| p{0.5cm} | p{5.5cm} | p{8cm} |} \hline
\textbf{Nr.} & \textbf{(executable) Software metadata description} & \textbf{Please fill in this column} \\ \hline
C1  & Current code version                         & 1.0 \\ \hline
C2  & Permanent link to code/repository used of this code version& https://github.com/kdis-lab/miml \\ \hline
C3  & Legal Code License                            & GPLv3 \\ \hline
C4  & Code versioning system used                   & git \\ \hline
C5  & Software code languages, tools, and services used & Java version 1.8\\ \hline
C6  & Compilation requirements, operating environments \& dependencies\& dependencies & junit 4.12\\ \hline 
C7  & If available Link to developer documentation/manual &  https://github.com/kdis-lab/MIML/blob/master/ documentation/MIML-UserManual.pdf \\ \hline
C8 & Support email for questions  & azafra@uco.es \\ \hline
\end{tabular}

\label{table:softwaremetadata2}
\end{table}

\end{document}